\documentclass{article}

\PassOptionsToPackage{numbers, compress}{natbib}
\usepackage{graphicx}
\usepackage{hyperref}
\usepackage[all]{hypcap} 

 \usepackage[dblblindworkshop, final]{neurips_2026}
\workshoptitle{Global South AI}

\usepackage[utf8]{inputenc} 
\usepackage[T1]{fontenc}    
\usepackage{hyperref}       
\usepackage{url}            
\usepackage{booktabs}       
\usepackage{amsfonts}       
\usepackage{nicefrac}       
\usepackage{microtype}      
\usepackage{xcolor}         

\title{The Vote Hides the Failure: Aggregation Choice and Noise Robustness in Heart Murmur Detection}

\author{%
Nicholaus Dismas Ladislaus \\
Carnegie Mellon University Africa\\
\texttt{eladisla@andrew.cmu.edu} \\
  \And
  Damilare Emmanuel  Olatunji \\
  Carnegie Mellon University Africa \\
\texttt{dolatunj@andrew.cmu.edu} \\
  \And
  Samuel Chol Buol \\
  Carnegie Mellon University Africa \\
\texttt{sbuol@andrew.cmu.edu} \\
}

\begin{document}

\maketitle

\begin{abstract}
Noise robustness in automated phonocardiogram (PCG) murmur detection, and how it is measured, remains underexamined despite growing interest in low-resource screening. We evaluate two independently reimplemented pipelines, Hierarchical Multi-Scale Convolutional Network (HMS-Net)---CNN, and Bidirectional Long Short-Term Memory (BiLSTM)---LSTM, under controlled, multi-severity noise with noise-augmented fine-tuning and held-out generalization testing. Under matched aggregation, the complete BiLSTM pipeline outperforms the complete HMS-Net pipeline across all conditions in accuracy and Weighted Accuracy. A stable aggregate accuracy score can misrepresent what individual predictions show: HMS-Net's native aggregation degrades under salt-and-pepper noise far less than majority-vote (MV) aggregation at the same severity, a gap reflecting window-level disagreement its native rule absorbs, while BiLSTM's MV accuracy rises after noise-augmented training even though its individual predictions do not improve. HMS-Net's training effect is significant under one accuracy metric but not another. Noise-robustness conclusions can depend as much on evaluation choices as on the models themselves.
\end{abstract}

\section{Introduction}
\label{introduction}

Public datasets collected under real clinical conditions, such as CirCor DigiScope~\cite{oliveiraCirCorDigiScopeDataset2022a, reynaHeartMurmurDetection2023}, have made Phonocardiogram (PCG)-based murmur detection increasingly promising alongside advances in deep learning, especially where a shortage of specialists limits auscultation-based diagnosis. PCG signals are vulnerable to acoustic perturbation from ambient noise, patient movement, loss of stethoscope contact, and the acquisition process itself. All of which degrades signal quality and may affect downstream classification~\cite{jakubecRobustDeepLearning2025}. Prior work investigates noise handling in PCG classification (Section~\ref{related}), but how the behaviour of declared best-performing pipelines compares under controlled acoustic perturbation, and whether evaluation conventions shape the conclusions drawn, remain open questions. We study this with a controlled, multi-severity noise-injection comparison of two independently reimplemented PCG pipelines, HMS-Net ~\cite{xuHierarchicalMultiScaleConvolutional2022} and
BiLSTM~\cite{monteiroDetectionHeartSound2022}, on CirCor DigiScope. Our contribution is an evaluation-focused stress test of whether robustness conclusions shift under aggregation rule, noise type, severity, and metric choice: comparing pipeline robustness under matched aggregation and three accuracy conventions, characterizing vulnerability across four noise types and severities, and testing generalization to a held-out noise type via noise-augmented fine-tuning. We find that a single aggregate accuracy number can hide failure in opposite directions: HMS-Net's patient-level accuracy looks stable while its window-level predictions disagree; BiLSTM's Majority-vote (MV) accuracy rises after training even though its individual predictions do not improve. This suggests that noise-robustness claims for PCG models -- and the evaluation choices behind them -- deserve the same scrutiny as the models themselves.

\section{Related works}
\label{related}

Prior comparative PCG studies test multiple architectures on CirCor DigiScope without a controlled noise protocol. Patwa et al.~\cite{patwaHeartMurmurAbnormal2025} compare 1D-CNN, LSTM, and convolutional-recurrent models, but remove noise-only segments and do not characterize robustness to noise; they use patient-level voting to improve accuracy, but do not examine it as a possible source of error. Aggregate metrics hiding degradation is an established concern: hidden stratification can mask poor performance on important subsets~\cite{oakden-raynerHiddenStratificationCauses2020}. Separately, adversarially-augmented training can raise error on clean data even as robustness improves~\cite{raghunathanUnderstandingMitigatingTradeoff2020a}, while ensemble and smoothing-based aggregation typically improves reported metrics~\cite{palUnderstandingNoiseAugmentedTraining2023}.
These findings have not been combined to test whether aggregation conceals training-induced regression. In PCG specifically, segment-to-recording aggregation is standard practice, but treated as engineering, not something to stress-test~\cite{patwaHeartMurmurAbnormal2025, neharyPhonocardiogramClassificationLearning2024}.
One segmentation study notes, as a limitation, that MV aggregation can miss infrequent pathological events~\cite{nomanMarkovSwitchingModelApproach2020} -- related to, but distinct from, the noise-induced masking we study systematically. Other PCG noise-robustness work reports degradation at the overall classification level~\cite{azamCardiacAnomalyDetection2022, shariatpanahExploringImpactNoise2023}, including evidence that accuracy depends on how noise is distributed across a recording, not just its severity~\cite{shariatpanahExploringImpactNoise2023}---without asking whether aggregation itself hides segment-level instability. To the best of our knowledge, no current PCG work isolates this pattern: how predictions are aggregated into a patient-level label can determine whether an accuracy measure stays stable, degrades, or improves---across two independently reimplemented pipelines under controlled noise.

\section{Methodology}
\label{methodology}

\subsection{Dataset and Architectures}

We use CirCor DigiScope~\cite{oliveiraCirCorDigiScopeDataset2022a} as released for the George B.\ Moody PhysioNet Challenge 2022~\cite{reynaHeartMurmurDetection2023}: 942 patients, 3046 recordings, Present 19.0\%/Unknown 7.2\%/Absent 73.8\%. We split this into 5 patient-level folds. We reimplement HMS-Net ~\cite{xuHierarchicalMultiScaleConvolutional2022} and a BiLSTM~\cite{monteiroDetectionHeartSound2022}, verified for structural and behavioral fidelity against their original submissions. Our clean baseline reaches 0.8021 $\pm$ 0.0228 (HMS-Net) and 0.7464 (BiLSTM) official \textit{\textit{W.acc}}, close to the originals' reported 0.81 and 0.757, respectively.

\subsection{Noise Injection and Aggregation}

We inject four noise types at inference -- Additive White Gaussian (AWGN), pink, and uniform at $\{20,10,5,0\}$~dB; and salt-and-pepper at $\{1,3,5,10\}\%$ (amplitude relative to local signal standard deviation)---on the raw waveform before each pipeline-specific preprocessing, which we found alters effective input signal-to-noise-ratio (SNR) by 5--12~dB in opposite directions per architecture, motivating full-pipeline evaluation. Each pipeline reports both \emph{native aggregation}---its own paper-specified rule, and \emph{MV}---a flat, architecture-neutral rule applied identically to both; neither corrects the other.

\subsection{ Training and Evaluation}

Both architectures are fine-tuned with per-example, randomized-severity noise under two variants -- full-range (uniform-ablation) and a range-truncated "vulnerability-informed" variant -- with salt-and-pepper — sparse and impulsive, unlike the additive continuous training noise — withheld as the held-out generalization test; this conflates unseen-noise and cross-family effects, separable via leave-one-noise-out in future work. Training protocol is fixed and mirrored identically across architectures for comparability. We report accuracy, Balanced Accuracy(unweighted mean of per-class recall), and Weighted Accuracy (\textit{\textit{W.acc}})~\cite{reynaHeartMurmurDetection2023}:
\[
s_{murmur} = \frac{5m_{PP}+3m_{UU}+m_{AA}}
{5(m_{PP}+m_{UP}+m_{AP})+3(m_{PU}+m_{UU}+m_{AU})+(m_{PA}+m_{UA}+m_{AA})}
\]
where $m_{XY}$ is the count of recordings with predicted label $X$ and true label $Y$ (Present, Unknown, Absent) -- which can diverge substantially from accuracy and Balanced Accuracy under class imbalance; effects are tested with paired per-fold $t$-tests (df=4), Bonferroni-corrected across 8 cells ($\alpha=0.00625$).

\section{Results and Discussion}
\label{results}
\begin{figure}[t]
  \centering
  \includegraphics[width=\linewidth]{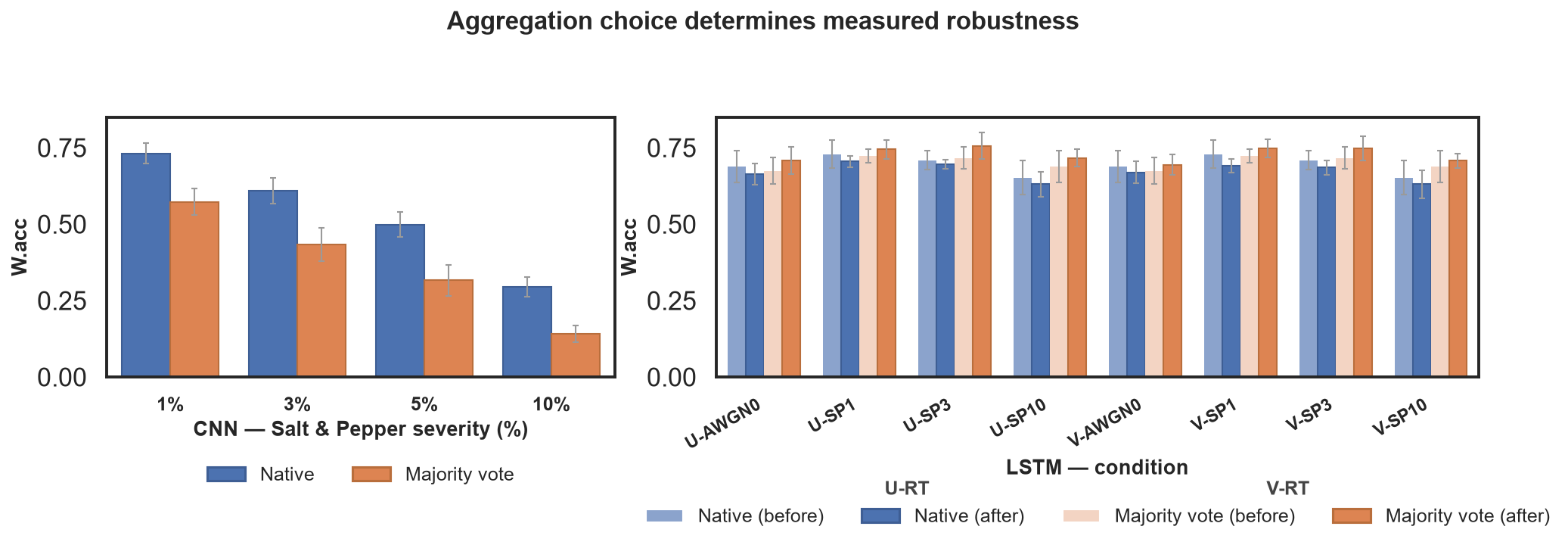}
  \caption{Aggregation choice determines measured robustness. Left: HMS-Net
native vs.\ MV W.acc under salt-and-pepper severity; native
accuracy exceeds MV in 59/60 cells, corresponding to
window-level disagreement that patient-level accuracy does not show.
Right: BiLSTM native and MV W.acc before/after noise-augmented 
training (U-RT: uniform-ablation; V-RT: vulnerability-informed);
MV accuracy improves while native accuracy does not. The
choice of aggregation changes which underlying prediction behaviour is visible.}
\label{fig:mask_bars}
\end{figure}

\begin{figure}[t]
  \centering
  \includegraphics[width=\linewidth]{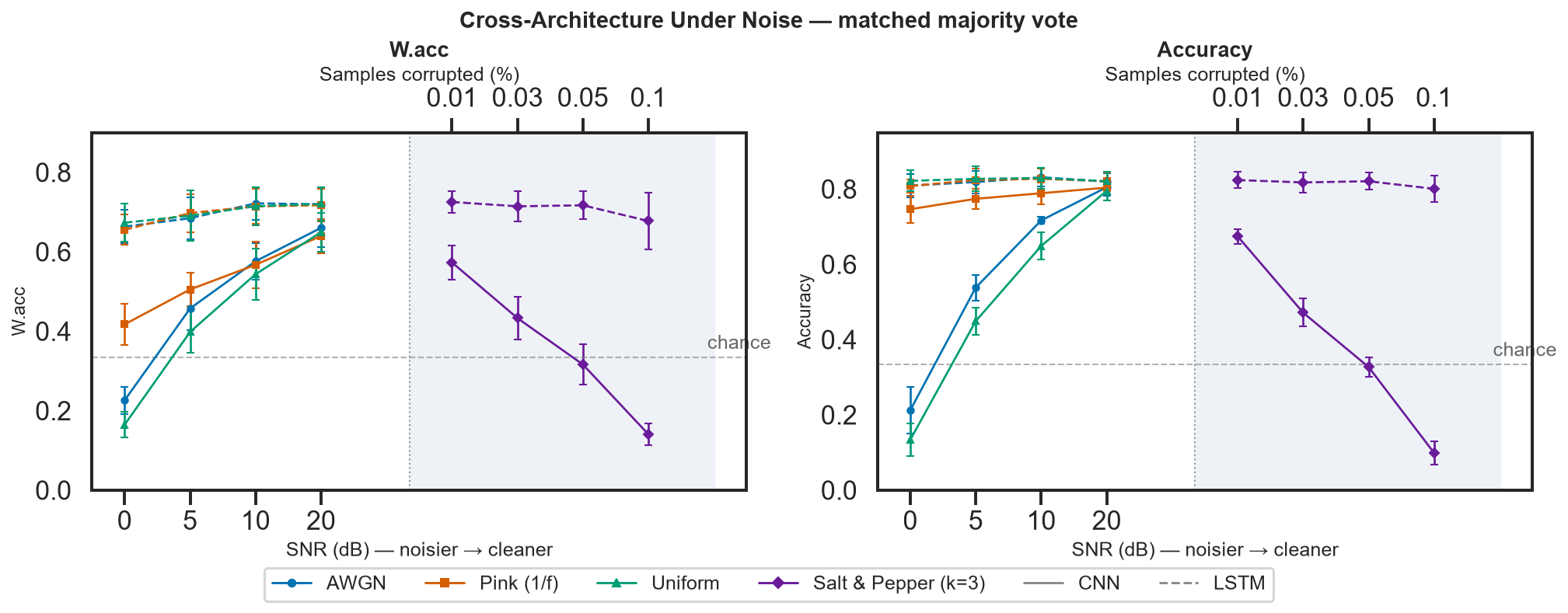}
  \caption{HMS-Net vs.\ BiLSTM, matched MV aggregation, all four
  noise types, \textit{W.acc} and Accuracy. BiLSTM leads 16/16 conditions under
  both metrics; under Balanced Accuracy (not shown), 13/16, with
  three mild-severity exceptions $<$1.1pp.}
  \label{fig:cross_arch}
\end{figure}

With each pipeline's own native aggregation, BiLSTM's accuracy stays within a 0.68--0.79 band across all four noise types, degrading only at the harshest salt-and-pepper corruption (0.683 at 10\%); HMS-Net collapses to 0.23 at 10\% salt-and-pepper and 0.30 at 0dB uniform noise, while exhibiting a similar characteristic under AWGN and uniform
noise (Appendix~\ref{app:vulnprofile}, Fig.~\ref{fig:cross_arch}). Because preprocessing alters effective input SNR differently (BiLSTM +5 to +12dB, HMS-Net -4 to -9dB relative to nominal), we report that the BiLSTM pipeline is more robust than the HMS-Net pipeline under this protocol; this comparison cannot isolate whether the difference stems from architecture, preprocessing, or their interaction. A matched-SNR comparison or a preprocessing-stage ablation would be needed to attribute the difference to architecture specifically, and we leave this to future work. Using matched MV aggregation across all four noise types (Fig.~\ref{fig:cross_arch}), BiLSTM outperforms HMS-Net in Accuracy and \textit{W.acc} at every condition all 16 conditions (based on descriptive comparison); under Balanced Accuracy, 13/16, with three mild-severity exceptions favoring HMS-Net by $<$1.1pp. The gap widens sharply at low SNR (HMS-Net Accuracy 0.212 vs.\ BiLSTM 0.809 at AWGN 0dB). We report two cases where a stable aggregate accuracy does not reflect what individual or window-level predictions show, via distinct mechanisms, on two pipelines (Fig.~\ref{fig:mask_bars}). For HMS-Net, native aggregation is Present-biased by construction (any-Present-wins at the patient-level), producing a native-vs-MV gap even without noise: on raw data, native exceeds MV by 2.9pp overall, from a +30.7pp Present-recall gain partly offset by a 19.1pp Unknown-recall loss. Under salt-and-pepper, this gap widens to 7.3pp, 16.8pp, and 13.4pp at 1\%, 5\%, and 10\% corruption, with native exceeding majority vote in 59 of 60 tested cells (98.3\%). The baseline 2.9pp reflects the aggregation rule's structural bias; CirCor provides no window-level ground truth, but injected severity cannot correlate with true murmur distribution, so the widening beyond baseline is noise-induced window-level disagreement absorbed by native aggregation's redundancy. For BiLSTM, after noise-augmented fine-tuning, majority-vote W.acc
improves at every condition (+2.1 to +5.3pp, two cells reaching
Bonferroni significance, Table ~\ref{tab:rt-stats}), while native
accuracy is flat or negative throughout ($-0.05$ to $-3.3$pp), across
all 8 cells and both variants. The sign is consistent across cells
(majority-vote 8/8, native 0/8; sign test $p \approx 0.008$ both ways)
--- stronger evidence than any single cell's significance given the
limited power of 5-fold tests. Fine-tuning shifts vote margins in a
direction majority-vote rewards without improving individual
predictions. Under standard accuracy HMS-Net's noise-augmented training significantly harms generalisation to mild, held-out salt-and-pepper noise ($-8.3$pp at 1\%, $p<0.001$) while helping at severe salt-and-pepper. However, under \textit{W.acc}, this collapses to $-0.74$pp ($p=0.37$, paired per-fold t-test df=4, Bonferroni-corrected, not significant). The reason is     class-specific: recall fell only in the Absent class (0.765$\to$0.630) while Present rose (0.737$\to$0.834); since \textit{W.acc} weights Absent lowest ($\times$1), this loss is nearly invisible under it despite the drop in Accuracy.  This discrepancy demonstrates that whether an effect is statistically significant can depend on the evaluation metric choice.

\begin{table}[t]
\caption{Noise-augmented training effects on native and MV W.acc, per
condition (unif.: uniform-ablation; vuln.: vulnerability-informed; paired per fold t-test $df=4$; Bonferroni $\alpha=0.00625$). Bold = significant.}
\label{tab:rt-stats}
\centering
\small
\begin{tabular}{ll rr rr rr rr}
\toprule
& & \multicolumn{4}{c}{HMS-Net (\textit{\textit{W.acc}})} & \multicolumn{4}{c}{BiLSTM (\textit{\textit{W.acc}})} \\
\cmidrule(lr){3-6}\cmidrule(lr){7-10}
& & \multicolumn{2}{c}{native} & \multicolumn{2}{c}{MV} & \multicolumn{2}{c}{native} & \multicolumn{2}{c}{MV} \\
\cmidrule(lr){3-4}\cmidrule(lr){5-6}\cmidrule(lr){7-8}\cmidrule(lr){9-10}
Var. & Cond. & $\Delta$ & $p$ & $\Delta$ & $p$ & $\Delta$ & $p$ & $\Delta$ & $p$ \\
\midrule
unif. & AWGN 0dB  & $+6.2$ & .122 & $+2.0$ & .504 & $-2.5$ & .312 & $\mathbf{+3.4}$ & $\mathbf{<.001}$ \\
unif. & S\&P 1\%  & $+0.5$ & .725 & $+4.5$ & .007 & $-2.4$ & .346 & $+2.1$ & .042 \\
unif. & S\&P 3\%  & $+6.0$ & .072 & $+4.0$ & .042 & $-1.3$ & .408 & $+3.9$ & .025 \\
unif. & S\&P 10\% & $+5.1$ & .144 & $+2.3$ & .261 & $-2.1$ & .550 & $+2.8$ & .286 \\
vuln. & AWGN 0dB  & $+6.1$ & .122 & $+1.3$ & .575 & $-1.8$ & .527 & $+2.0$ & .056 \\
vuln. & S\&P 1\%  & $-0.7$ & .370 & $+6.0$ & .008 & $-3.7$ & .187 & $+2.4$ & .048 \\
vuln. & S\&P 3\%  & $+5.6$ & .066 & $+5.1$ & .009 & $-2.5$ & .237 & $\mathbf{+3.1}$ & $\mathbf{<.001}$ \\
vuln. & S\&P 10\% & $+4.6$ & .136 & $+1.9$ & .347 & $-2.1$ & .628 & $+1.9$ & .392 \\
\bottomrule
\end{tabular}
\end{table}

\section{Limitations and Conclusion}
\label{limits}

Several items remain open. Per-recording predicted-class distributions across severity are not reported here; this is part of a forthcoming extension. Both pipelines' native aggregation is present-biased by construction, contributing a baseline gap independent of noise; we report the gap's growth with severity instead of its absolute value (Section~\ref{results}). Noise-augmented fine-tuning preserves BiLSTM's clean-data accuracy: MV accuracy remains above 0.76 across all five folds and both variants. An ablation reducing recordings per patient, testing whether HMS-Net's noise robustness depends on redundancy unavailable in single-visit screening, is deferred. Per-fold seed variance cannot be separated from fold variance, since each fold used one seed. Unknown-class recall estimates rest on 10--16 patients per fold and should be read as noisy. All injected noise is synthetic and may not transfer to real-world interference.

\begin{ack}
The authors thank the AI Healthcare Lab at Carnegie Mellon University
Africa, led by Prof. Carine P. Mukamakuza, and Research Associates
Godbright Uiso, Julius Zannu, and Mona Aman, for their assistance with
this ongoing work, which began during a summer internship in the Lab's
Cardiovascular Diseases section.

This work received no external funding. The authors declare no
competing interests.
\end{ack}

\begingroup
\small
\bibliographystyle{unsrtnat}
\bibliography{references}

@article{reynaHeartMurmurDetection2023,
	title = {Heart murmur detection from phonocardiogram recordings: {The} {George} {B}. {Moody} {PhysioNet} {Challenge} 2022},
	volume = {2},
	issn = {2767-3170},
	shorttitle = {Heart murmur detection from phonocardiogram recordings},
	url = {https://dx.plos.org/10.1371/journal.pdig.0000324},
	doi = {10.1371/journal.pdig.0000324},
	language = {en},
	number = {9},
	urldate = {2026-08-31},
	journal = {PLOS Digital Health},
	author = {Reyna, Matthew A. and Kiarashi, Yashar and Elola, Andoni and Oliveira, Jorge and Renna, Francesco and Gu, Annie and Perez Alday, Erick A. and Sadr, Nadi and Sharma, Ashish and Kpodonu, Jacques and Mattos, Sandra and Coimbra, Miguel T. and Sameni, Reza and Rad, Ali Bahrami and Clifford, Gari D.},
	editor = {Celi, Leo Anthony},
	month = sep,
	year = {2023},
	pages = {e0000324},
}

@article{shariatpanahExploringImpactNoise2023,
	title = {Exploring the impact of noise and degradations on heart sound classification models},
	volume = {85},
	issn = {17468094},
	url = {https://linkinghub.elsevier.com/retrieve/pii/S1746809423003658},
	doi = {10.1016/j.bspc.2023.104932},
	language = {en},
	urldate = {2026-08-29},
	journal = {Biomedical Signal Processing and Control},
	author = {Shariat Panah, Davoud and Hines, Andrew and McKeever, Susan},
	month = aug,
	year = {2023},
	pages = {104932},
}

@inproceedings{raghunathanUnderstandingMitigatingTradeoff2020a,
	title = {Understanding and {Mitigating} the {Tradeoff} between {Robustness} and {Accuracy}},
	issn = {2640-3498},
	url = {https://proceedings.mlr.press/v119/raghunathan20a.html},
	language = {en},
	urldate = {2026-08-29},
	booktitle = {Proceedings of the 37th {International} {Conference} on {Machine} {Learning}},
	publisher = {PMLR},
	author = {Raghunathan, Aditi and Xie, Sang Michael and Yang, Fanny and Duchi, John and Liang, Percy},
	month = nov,
	year = {2020},
	pages = {7909--7919},
}

@inproceedings{oakden-raynerHiddenStratificationCauses2020,
	address = {New York, NY, USA},
	series = {{CHIL} '20},
	title = {Hidden stratification causes clinically meaningful failures in machine learning for medical imaging},
	isbn = {978-1-4503-7046-2},
	url = {https://dl.acm.org/doi/10.1145/3368555.3384468},
	doi = {10.1145/3368555.3384468},
	urldate = {2026-08-29},
	booktitle = {Proceedings of the {ACM} {Conference} on {Health}, {Inference}, and {Learning}},
	publisher = {Association for Computing Machinery},
	author = {Oakden-Rayner, Luke and Dunnmon, Jared and Carneiro, Gustavo and Re, Christopher},
	month = apr,
	year = {2020},
	pages = {151--159},
}

@article{azamCardiacAnomalyDetection2022,
	title = {Cardiac anomaly detection considering an additive noise and convolutional distortion model of heart sound recordings},
	volume = {133},
	issn = {09333657},
	url = {https://linkinghub.elsevier.com/retrieve/pii/S0933365722001695},
	doi = {10.1016/j.artmed.2022.102417},
	language = {en},
	urldate = {2026-08-28},
	journal = {Artificial Intelligence in Medicine},
	author = {Azam, Farhat Binte and Ansari, Md. Istiaq and Nuhash, Shoyad Ibn Sabur Khan and McLane, Ian and Hasan, Taufiq},
	month = nov,
	year = {2022},
	pages = {102417},
}

@article{nomanMarkovSwitchingModelApproach2020,
	title = {A {Markov}-{Switching} {Model} {Approach} to {Heart} {Sound} {Segmentation} and {Classification}},
	volume = {24},
	issn = {2168-2208},
	url = {https://ieeexplore.ieee.org/document/8746548/},
	doi = {10.1109/JBHI.2019.2925036},
	number = {3},
	urldate = {2026-08-27},
	journal = {IEEE Journal of Biomedical and Health Informatics},
	author = {Noman, Fuad and Salleh, Sh-Hussain and Ting, Chee-Ming and Samdin, S. Balqis and Ombao, Hernando and Hussain, Hadri},
	month = mar,
	year = {2020},
	pages = {705--716},
}

@article{neharyPhonocardiogramClassificationLearning2024,
	title = {Phonocardiogram {Classification} by {Learning} {From} {Positive} and {Unlabeled} {Examples}},
	volume = {73},
	issn = {1557-9662},
	url = {https://ieeexplore.ieee.org/document/10456924/},
	doi = {10.1109/TIM.2024.3372221},
	urldate = {2026-08-27},
	journal = {IEEE Transactions on Instrumentation and Measurement},
	author = {Nehary, Ebrahim A. and Rajan, Sreeraman},
	year = {2024},
	pages = {1--14},
}

@misc{palUnderstandingNoiseAugmentedTraining2023,
	title = {Understanding {Noise}-{Augmented} {Training} for {Randomized} {Smoothing}},
	url = {http://arxiv.org/abs/2305.04746},
	doi = {10.48550/arXiv.2305.04746},
	urldate = {2026-08-27},
	publisher = {arXiv},
	author = {Pal, Ambar and Sulam, Jeremias},
	month = may,
	year = {2023},
	note = {arXiv:2305.04746 [cs.LG]},
}

@article{patwaHeartMurmurAbnormal2025,
	title = {Heart {Murmur} and {Abnormal} {PCG} {Detection} via {Wavelet} {Scattering} {Transform} and {1D}-{CNN}},
	volume = {25},
	issn = {1558-1748},
	url = {https://ieeexplore.ieee.org/document/10899755/},
	doi = {10.1109/JSEN.2025.3541320},
	number = {7},
	urldate = {2026-08-27},
	journal = {IEEE Sensors Journal},
	author = {Patwa, Ahmed and Mahboob Ur Rahman, Muhammad and Al-Naffouri, Tareq Y.},
	month = apr,
	year = {2025},
	pages = {12430--12443},
}

@article{jakubecRobustDeepLearning2025,
	title = {A robust deep learning based model for denoising phonocardiogram signals in clinical environments},
	volume = {161},
	issn = {09521976},
	url = {https://linkinghub.elsevier.com/retrieve/pii/S0952197625022948},
	doi = {10.1016/j.engappai.2025.112286},
	language = {en},
	urldate = {2026-08-23},
	journal = {Engineering Applications of Artificial Intelligence},
	author = {Jakubec, Maros and Lieskovska, Eva and Pocta, Peter},
	month = dec,
	year = {2025},
	pages = {112286},
}

@article{oliveiraCirCorDigiScopeDataset2022a,
	title = {The {CirCor} {DigiScope} {Dataset}: {From} {Murmur} {Detection} to {Murmur} {Classification}},
	volume = {26},
	copyright = {https://creativecommons.org/licenses/by/4.0/legalcode},
	issn = {2168-2194, 2168-2208},
	shorttitle = {The {CirCor} {DigiScope} {Dataset}},
	url = {https://ieeexplore.ieee.org/document/9658215/},
	doi = {10.1109/JBHI.2021.3137048},
	number = {6},
	urldate = {2026-08-23},
	journal = {IEEE Journal of Biomedical and Health Informatics},
	author = {Oliveira, Jorge and Renna, Francesco and Costa, Paulo Dias and Nogueira, Marcelo and Oliveira, Cristina and Ferreira, Carlos and Jorge, Alipio and Mattos, Sandra and Hatem, Thamine and Tavares, Thiago and Elola, Andoni and Rad, Ali Bahrami and Sameni, Reza and Clifford, Gari D. and Coimbra, Miguel T.},
	month = jun,
	year = {2022},
	pages = {2524--2535},
}

@inproceedings{monteiroDetectionHeartSound2022,
	title = {Detection of {Heart} {Sound} {Murmurs} and {Clinical} {Outcome} with {Bidirectional} {Long} {Short}-{Term} {Memory} {Networks}},
	volume = {498},
	issn = {2325-887X},
	url = {https://ieeexplore.ieee.org/document/10081753},
	doi = {https://doi.org/10.22489/CinC.2022.153},
	urldate = {2026-07-07},
	booktitle = {2022 {Computing} in {Cardiology} ({CinC})},
	author = {Monteiro, Sofia and Fred, Ana and da Silva, Hugo Plácido},
	month = sep,
	year = {2022},
	note = {ISSN: 2325-887X},
	pages = {1--4},
}

@inproceedings{xuHierarchicalMultiScaleConvolutional2022,
	title = {Hierarchical {Multi}-{Scale} {Convolutional} {Network} for {Murmurs} {Detection} on {PCG} {Signals}},
	volume = {498},
	issn = {2325-887X},
	url = {https://ieeexplore.ieee.org/document/10081900},
	doi = {10.22489/CinC.2022.439},
	urldate = {2026-07-07},
	booktitle = {2022 {Computing} in {Cardiology} ({CinC})},
	author = {Xu, Yujia and Bao, Xinqi and Lam, Hak-Keung and Kamavuako, Ernest N.},
	month = sep,
	year = {2022},
	note = {ISSN: 2325-887X},
	pages = {1--4},
}
\endgroup


\appendix

\section{Dataset Detail}
\label{app:dataset}
CirCor DigiScope, as released for the PhysioNet 2022 Challenge: 942 patients, 3163 recordings, reduced to 3046 after excluding non-murmur-location recordings for Present-class patients (following both pipelines' original preprocessing). Class distribution: Present 179 (19.0\%), Unknown 68 (7.2\%), Absent 695 (73.8\%). Patients are assigned to 5 folds by sorting patient IDs lexicographically and applying round-robin assignment ($i \bmod 5$), identically for both pipelines. Per-fold patient counts: 189, 189, 188, 188, 188.
Per-fold class counts: (42,14,133), (42,10,137), (34,16,138), (25,13,150), (36,15,137) for Present/Unknown/Absent respectively. We confirmed no patient's recordings are split across folds (leakage check, both pipelines). Recordings per patient range from 1 to 6 (mean 3.36, median 4; distribution: 62 patients with 1 recording, 156 with 2, 123 with 3, 588 with 4, 10 with 5, 3 with 6).

\section{Pipeline Fidelity}
\label{app:fidelity}

HMS-Net was verified via full forward-pass shape tracing against the original architecture diagram, state-dict key correspondence, and bitwise-identical checkpoint round-trip. The original architecture contains no native dropout; we add \texttt{nn.Dropout} between global average pooling and the final fully-connected layer (default rate 0, active only for MC Dropout checkpoints), leaving the clean baseline architecturally identical to the original. BiLSTM was verified via line-for-line reproduction of the reference model-building code and layer-class/parameter-count equality against the original TensorFlow implementation; the original splits patches via a patch-level train/test split, while our reimplementation splits by patient to prevent leakage, a deliberate, more conservative deviation.

\paragraph{Native aggregation rules.} 

For HMS-Net, per-window softmax outputs are converted to per-class time durations at the recording level; a recording is labeled Unknown if the Unknown-duration fraction exceeds 0.8, Present if Present duration reaches at least 3.0 seconds, and Absent otherwise. At the patient level, any Present recording yields a Present label; failing that, any Unknown recording (with fewer than two Absent recordings) yields Unknown; otherwise Absent. For BiLSTM, per-recording patch probabilities are mean-pooled to one vector per recording; the recording with the highest P(Present) is selected, and its argmax becomes the patient label. Both rules are Present-biased by construction (any-Present-wins for HMS-Net, max-P (Present) selection for BiLSTM), which contributes a baseline native-vs-MV gap independent of noise (Section~\ref{results}).

\section{Noise Parameterization}
\label{app:noise}

Salt-and-pepper corruption amplitude is set relative to local signal standard deviation ($k=3$), as opposed to full dynamic range. An initial full-range parameterization was found to cause catastrophic accuracy collapse even at 1\% corruption, a result of corrupted-sample magnitude and not model fragility; $k=3$ local-std scaling corrects this while preserving a meaningful corruption signal.

\section{Noise-Augmented Fine-Tuning: Full Protocol}
\label{app:rt}

Per training example, noise type is drawn uniformly at random from {AWGN, pink, uniform} and severity is sampled continuously and uniformly at random within the active variant’s range; every example receives noise (no clean passthrough). Uniform-ablation: full {0, 20} dB range for all three types. Vulnerability-informed: range truncated to {0, 10} dB for AWGN and uniform (pink unchanged at {0, 20} dB) -- this truncation is derived from HMS-Net’s vulnerability profile and applied identically to both pipelines; for BiLSTM, whose vulnerability profile differs, the label "vulnerability-informed" describes the training protocol’s origin, not a BiLSTM-specific optimization. Salt-and-pepper is withheld entirely from fine-tuning. Both variants: 20 epochs, AdamW (learning rate 10-4, weight decay 0), label smoothing 0.1, batch size 128, ReduceLROnPlateau (factor 0.1, patience 5) on training loss, full-model fine-tuning from each pipeline’s designated baseline checkpoint (deterministic for HMS-Net; MC-Dropout for BiLSTM), base seed 42 with fold seed = 42 + k.

\section{Vulnerability Profiles (Extended)}
\label{app:vulnprofile}

\begin{figure}[h]
  \centering
  \includegraphics[width=\linewidth]{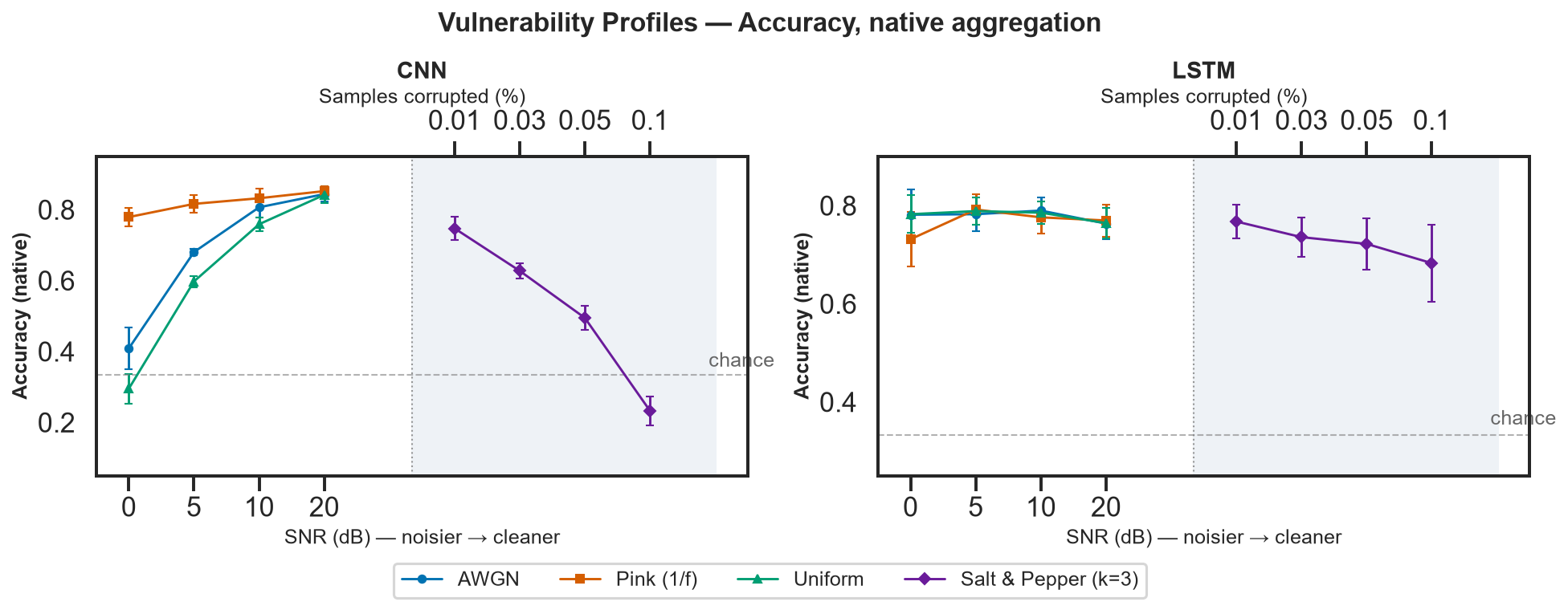}
  \caption{HMS-Net and BiLSTM native-aggregation accuracy under four noise
  types. BiLSTM remains within a 0.68--0.79 band across nearly all
  conditions; HMS-Net degrades sharply under harsh salt-and-pepper and
  cannot distinguish AWGN from uniform noise at any tested severity.}
  \label{fig:vulnprofile}
\end{figure}

Fig.~\ref{fig:vulnprofile} shows each pipeline's accuracy under its own native aggregation, across all four noise types, before any cross-pipeline comparison or aggregation-choice analysis. We include it as supporting context, because Fig.~\ref{fig:cross_arch} already establishes the cross-pipeline comparison under matched conditions, and Fig.~\ref{fig:mask_bars} already carries the paper's central claims; this figure's role is orientation, not evidence the main argument depends on.

Two distinct characteristics can be observed. First, BiLSTM's native accuracy is comparatively insensitive to noise type or severity, remaining within a 0.68--0.79 band across nearly all conditions and degrading meaningfully only under the harshest salt-and-pepper corruption (0.683 at 10\%). HMS-Net shows no such stability: Accuracy falls to 0.23
under 10\% salt-and-pepper and to 0.30 under 0dB uniform noise, and shows similar Accuracy under AWGN and uniform noise at every tested severity (mean difference $\leq$4.2 percentage points). This early, single-pipeline-convention view corresponds to the matched comparison in Section~\ref{results} and the redundancy-masking mechanism identified in Fig.~\ref{fig:mask_bars}a, though it is not itself sufficient to establish either claim, since the two panels here use each pipeline's own aggregation rule.

\section{Uncertainty Quantification}
\label{app:uq}

We compare Monte Carlo Dropout (MCD, N = 10) against Deep Ensembles (4-of-5 folds) via Expected Calibration Error (ECE), for both pipelines, across all four noise types and severities, on non-reinforced (baseline) checkpoints. The two pipelines disagree on which method is better calibrated. For HMS-Net, Deep Ensembles are consistently better calibrated than MCD (mean ECE 0.314 vs. 0.328 across 80 evaluated units); for BiLSTM, MCD is consistently better than Ensembles (0.210 vs. 0.222). Neither pipeline shows the other’s ordering at any tested noise type or severity.

We had intended to test whether noise-augmented training changes this calibration gap. This comparison is not possible with current results: ECE or selective-accuracy metrics on the Noise-augmented-trained checkpoints are yet to be computed. The reinforced BiLSTM checkpoints retain the MC-Dropout layer inherited from their baseline checkpoint, whereas the reinforced HMS-Net checkpoints were trained from a deterministic baseline and contain no dropout. Completing the post-noise-augmented training comparison therefore requires new inference on the existing Noise-augmented-trained checkpoints and, for MCD specifically, a new dropout fine-tuning of the reinforced HMS-Net checkpoints. These analyses are in progress and therefore not reported here on incomplete grounds.


\newpage

\end{document}